\def\year{2022}\relax
\documentclass[letterpaper]{article} 
\usepackage{aaai22}  
\usepackage{times}  
\usepackage{helvet}  
\usepackage{courier}  
\usepackage[hyphens]{url}  
\usepackage{graphicx} 
\usepackage{natbib}  
\usepackage{caption} 
\DeclareCaptionStyle{ruled}{labelfont=normalfont,labelsep=colon,strut=off} 
\usepackage{algorithm}
\usepackage{algorithmic}
\usepackage{multicol}
\usepackage{multirow}
\usepackage{amssymb}
\usepackage{amsmath}
\usepackage{xcolor}

\usepackage{comment}

\usepackage{newfloat}
\usepackage{listings}
\floatstyle{ruled}
\newfloat{listing}{tb}{lst}{}
\floatname{listing}{Listing}
\title{\centering{AXM-Net: Implicit Cross-Modal Feature Alignment for Person Re-identification}}

\author {
    Ammarah Farooq\textsuperscript{\rm 1},
    Muhammad Awais\textsuperscript{\rm 1,2,3},
    Josef Kittler\textsuperscript{\rm 1,2,3},
    Syed Safwan Khalid\textsuperscript{\rm 1} 
}
\affiliations {
    \textsuperscript{\rm 1}Centre for Vision, Speech and Signal Processing (CVSSP), University of Surrey,\\
    \textsuperscript{\rm 2}Surrey Institute for People-centred AI (SI-PAI),\\ \textsuperscript{\rm 3}Sensus Futuris Ltd.\\
    \{ammarah.farooq, m.a.rana,  j.kittler,  s.khalid\}@surrey.ac.uk
}

\begin{document}

\maketitle

\begin{abstract}
Cross-modal person re-identification (Re-ID) is critical for modern video surveillance systems. The key challenge is to align cross-modality representations induced by the semantic information present for a person and ignore background information. This work presents a novel convolutional neural network (CNN) based architecture designed to learn semantically aligned cross-modal visual and textual representations. The underlying building block, named AXM-Block, is a unified multi-layer network that dynamically exploits the multi-scale knowledge from both modalities and re-calibrates each modality according to shared semantics. To complement the convolutional design, contextual attention is applied in the text branch to manipulate long-term dependencies. Moreover, we propose a unique design to enhance visual part-based feature coherence and locality information. Our framework is novel in its ability to implicitly learn aligned semantics between modalities during the feature learning stage. The unified feature learning effectively utilizes textual data as a super-annotation signal for visual representation learning and automatically rejects irrelevant information. The entire AXM-Net is trained end-to-end on CUHK-PEDES data. We report results on two tasks, person search and cross-modal Re-ID. The AXM-Net outperforms the current state-of-the-art (SOTA) methods and achieves 64.44\% Rank@1 on the CUHK-PEDES test set. It also outperforms its competitors by $>$10\% in cross-viewpoint text-to-image Re-ID scenarios on CrossRe-ID and CUHK-SYSU datasets.
\end{abstract}

\section{Introduction}
Person re-identification (Re-ID) has become a principal component of intelligent video surveillance systems that aims to retrieve a queried person from a large database of pedestrian images. The database typically contains non-overlapping camera viewpoints with respect to the query images. Depending on the type of information provided as a query, the task is referred to as person Re-ID or person search in the literature. Person search~\cite{li2017person} aims to find a person based on the natural language description of the person while images are provided as a query in person Re-ID. In person search, there is no constraint on the camera viewpoints of the person, i.e., in the visual gallery person may have the same pose for which the description is given. Nevertheless, in both tasks, it is critical to learn discriminative feature representations which are unique to an individual and well-aligned within the class for finer matching.  
 
 \begin{figure}
\centering
        \includegraphics[width=\linewidth, height=0.5\linewidth]{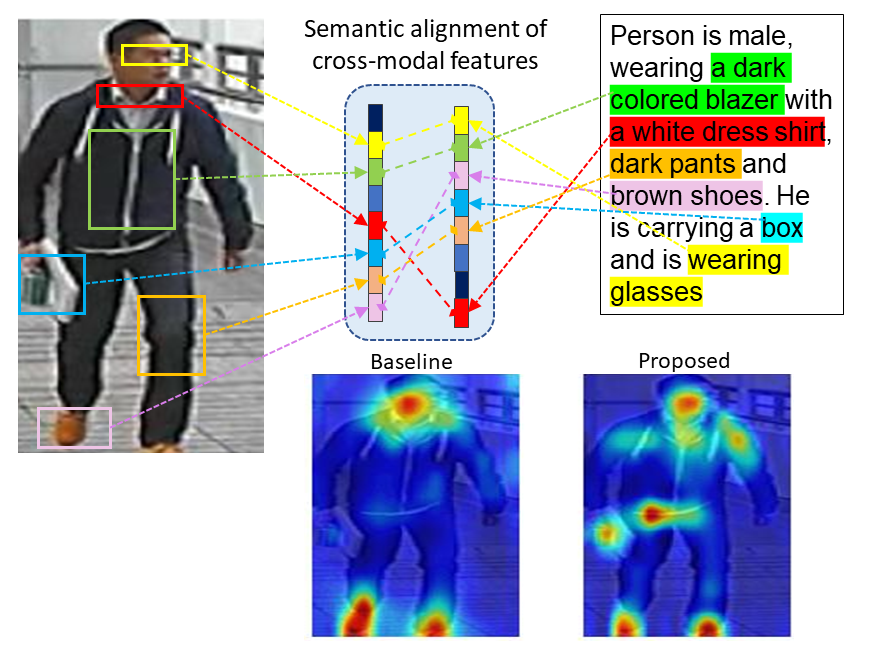}
        \footnotesize{\caption{Illustration of semantic alignment for visual and textual features. The semantic information in the features should be aligned to enhance the cross-modal associations.}
    \label{axm_intro}}
\end{figure}
 
 The literature is packed with numerous deep learning based person re-identification approaches. These methods aim to learn robust person representations~\cite{zhou2019omni,dai2019batch}, apply various attention mechanisms~\cite{Xia_2019_ICCV,Chen_2019_ICCV}, look for cross-domain knowledge transfer~\cite{Jing_2020_CVPR,Chen_2019_ICCV_cross} and so on. Cross-modal person Re-ID is another important aspect of Re-ID task~\cite{yan2018person,farooq2020convolutional,farooq2020IJCB,Lu_2020_CVPR,Jing_2020_CVPR}. The dependency on available image queries limits the practical application of a vision-based system. For example, in the case of criminal search, often the CCTV footage (or image) of a criminal is not available. Therefore, police rely on the unique cues of the criminal from the witnesses’ descriptions often given in terms of natural language description. In such cases, with no images available, this descriptive information serves as a query for person Re-ID. Hence, employing a multi-modal Re-ID system can overcome the limitations of image-based systems. 
 
 This work focuses on cross-modal person Re-ID using visual and textual information of the person. The aim of the work is to design a system that semantically aligns cross-modal and cross-pose representations implicitly by focusing on the information present in the two modalities. By implicit alignment we mean alignment of two modalities without using external cues, like segmentation, human body landmarks, attributes prediction from image etc. Note, that existing methods~\cite{aggarwal2020text,wang2020vitaa,jing2018pose} rely on complex external cues and explicit alignment of the feature embeddings. 
 There are several challenges while dealing with two distinct modalities. First, the structure of appearance information in both modalities is quite different. Presumably, images have persons always standing upright while the person description can have any sequence. Second, it is critical to learn a network that can extract the semantics in data instead of memorising corresponding image-text pairs for the identities seen during training. Third, the attributes information present in the features, for example, colour and type of clothes person is wearing, the activity of the person, accessories, etc, should be aligned across the modalities to learn the associations among image parts and textual phrases and disregard background noise (Figure~\ref{axm_intro}). Henceforth, the terms `semantic concepts’ and  `attributes’ of the person are used interchangeably.
 
The main idea of this work is to align the visual and textual features of a person to enable cross-modal search seamlessly. To achieve this goal, we present AXM-Net, a novel convolutional neural network (CNN) based architecture designed to deal with the challenges mentioned above and capable of learning semantically aligned cross-modal feature representations. The underlying building block consists of multiple streams of feature maps capturing a variable amount of intra-modal information and an alignment network that is trained based on the semantics present in both modalities. The output representations are, hence, attended by the fused cross-modal details. The alignment network leverages multi-context intra-modality information and cross-modal attributes to boost informative concepts and suppress the noisy or background information.
 
 Apart from exploiting inter-modal semantic knowledge, we also propose modality-specific contextual attention mechanisms to effectively extract within-modal relevant information. To be specific, we introduce a visual part-based feature coherence module to locally enhance or suppress the features. We also note that a semantic attribute can be shared across parts, for example, long maxi dress covers upper body as well lower body of a person etc. Therefore, we also focus on consistency of a feature across parts. Since the unified feature learning is performed using a CNN backbone, the sequential nature of the language modality demands learning long-term associations among the person attributes. For example, a description may contain information about the head (hairs, style) at the beginning, followed by a description for the lower body, and again carry information about the head (wearing hat). Hence, we also employ contextual attention to learn these useful associations among the textual attributes.
 Our contributions can be summarised as follows:
 \begin{itemize}
     \item Unified cross-modal semantic alignment block (AXM-Block) is proposed to mutually learn representations based on person attributes. To our knowledge, this is the first work to employ implicit semantic alignment across modalities at feature learning stage in the Re-ID setting.
     
     \item We put forward effective image parts-based feature coherency learning mechanism. The proposed module attends local spatial region based details while exploiting context from other parts of the image.
     
     \item Extensive experiments demonstrate the superiority 
     of the proposed AXM-Net over the current SOTA by 3\% on the CUHK-PEDES~\cite{li2017person} benchmark and by a wide margin on CrossRe-ID and CUHK-SYSU~\cite{farooq2020convolutional,farooq2020IJCB} in all retrieval scenarios.
 \end{itemize}

\section{Related Work}
The task of person search by natural language descriptions was introduced by ~\cite{li2017person}. The proposed method was a CNN-RNN network to learn global level cross-modal features. The following works~\cite{li2017identity,chen2018Pwm} also focused on similar network architectures and a little improvement was observed in performance. Major improvements were shown by ~\cite{chen2018improving,zhang2018deep,zheng2020dual} where researchers start exploiting global-local associations and improving the feature embedding space. More recent works  ~\cite{jing2020pose,wang2020img,wang2020vitaa,aggarwal2020text} started employing auxiliary learning branches to explicitly make use of pose key-points, person attributes, segmentation masks, body parts and textual phrases. These approaches brought improvements as compared to using only global features. Zheng et. al.~\cite{gumbel2020} proposed to use Gumbel attention to extract strongly aligned features by performing global, phrase and word-level matchings across the image and sentence. The SOTA work~\cite{gao2021contextual} also used multi-scale features by employing a staircase network to extract the visual features at global, regional and patch level, and then aligned these features with textual feature by applying cross-modal non-local attention. 
As mentioned earlier, cross-modal searches help to overcome limitations of vision only systems. For text based person Re-ID,~\cite{yan2018person} published the pioneering work and reported results on the CUHK03 and Viper datasets under multiple retrieval scenarios. Recent works~\cite{farooq2020convolutional,farooq2020IJCB} proposed to jointly optimise the two modalities and applied canonical correlation analysis to enhance similarity between the corresponding features. These works also reported results for larger cross-modal test splits including CrossRe-ID and CUHK-SYSU.

All the above mentioned works extract visual features and textual features from separate backbones and then perform cross-modal alignment which mostly requires auxiliary tasks to increase matching performance. However, the proposed method emphasises on leveraging the multi-level representations with the aim of latent alignment of the cross-modal features during feature extraction stage.

\begin{figure*}
\centering
        \includegraphics[width=\textwidth,height=0.44\linewidth]{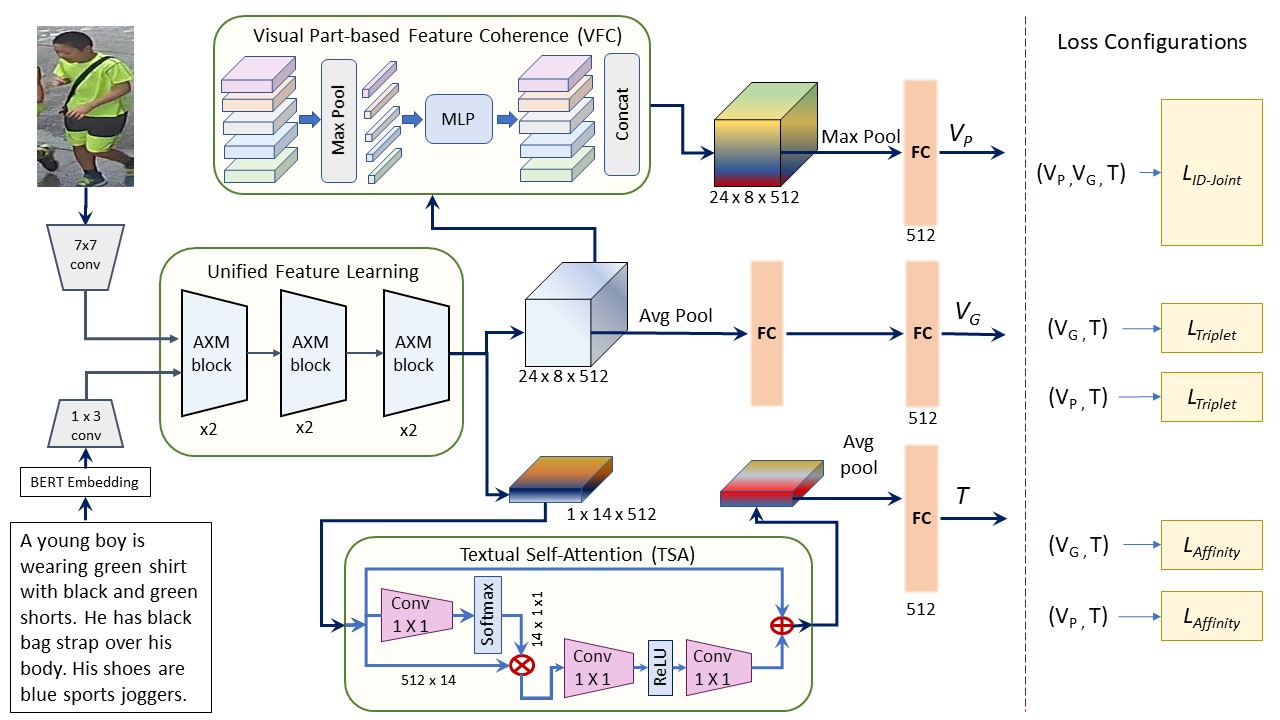}
        \footnotesize{\caption{Illustrative diagram of our cross-modal AXM-Net, which generates global visual feature $V_G$, part based visual feature $V_P$ and textual feature $T$. Softmax loss $\mathcal{L}_{ID_{joint}}$ is a function of all the features. Matching losses are trained pairwise with the textual feature for each visual feature.}
    \label{block-diagram}}
\end{figure*}

\section{Cross-modal AXM-Net Framework}
Figure~\ref{block-diagram} shows the block diagram of the proposed AXM-Net, which includes a unified feature learning network, visual part-based feature coherence (VFC) learning branch and global vision and textual branches. The details of each part are presented in the following subsections.
\subsection{Architecture of the AXM-Block} 
\begin{figure}
\centering
        \includegraphics[width=\linewidth,height=0.6\linewidth]{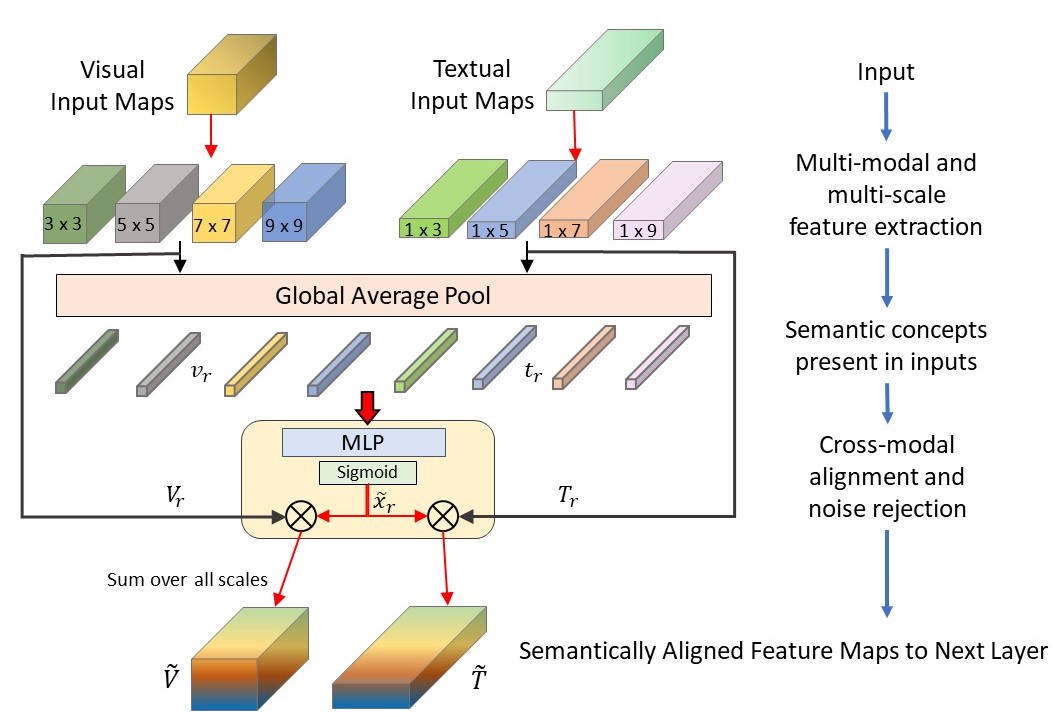}
        \footnotesize{\caption{AXM-Block design. Unified cross-modal feature learning block. $3 \times 3$ indicates kernel size of the convolution.}
        \label{axm_block}}
\end{figure}

First, we present the design of the proposed \textbf{cross-modal semantic alignment block} called as AXM-block shown in Figure~\ref{axm_block}. The design principle of the block is to exploit multi-scale and multi-modal information to align semantic information between the modalities. We define $X_r=(V_r,T_r)$ to be the visual and textual features maps at convolution filter scales $r =3,5,7,9$. We apply a global average pool operation to collect the global concepts present in each feature map generating $\mathbf{x_r=(v_r,t_r)}$ vectors of length spanning the entire channel dimension. These vectors are then passed to the alignment network together with the input feature maps.
\begin{equation}
    \mathbf{\tilde{x_r}} = \sigma(MLP(\mathbf{x_r}))
\end{equation}

The alignment network consists of a two-layer MLP with a sigmoid activation ($\sigma$) at the end to produce a scaling vector $\mathbf{{\tilde{x_r}=(\tilde{v_r},\tilde{t_r})}}$ corresponding to each feature vector. These scaling vectors convey the importance of each channel with respect to the shared information between modalities across different scales. Feature alignment is performed by multiplying each scaling vector by the corresponding input feature map. Finally, all attended multi-scale features are fused by summation and passed to the next layer.
\begin{equation}
    (\tilde{V}, \tilde{T}) = \sum_{r=3,5,7,9} (V_r \odot \mathbf{\tilde{v_r}},\  T_r \odot \mathbf{\tilde{t_r}} )
\end{equation}

\subsection{Unified Feature Learning using AXM-Block}
Intrinsically, the semantic information present in both modalities is the same, as both are describing the same person. However, the corresponding semantic concepts and their relationships may be available at different scales and locations within each modality. We propose a novel idea to align these representations across modalities based on the semantic information during feature extraction phase. For this purpose, we propose to use the AXM-Block as the basic building block of our feature learning backbone. The unified feature learning backbone consists of stacked AXM-blocks. Intra-modal multi-scale learning captures the locally critical context with respect to a spatial location. In contrast, inter-modal multi-scale alignment dynamically drives the two modalities to conform to each other. Each stream of channels contains a coarser-to-finer level of semantics present in the input image or text. Likewise, cross-modal multi-scale learning captures and semantically aligns an unaligned coarser-to-finer level of semantics present in the multi-modal input. It is important to understand the cross-modal unified feature learning for semantic alignment to differentiate from typical multi-scale feature learning methods~\cite{zhou2019omni,chen2017person,cai2019multi}.To illustrate this difference, we show baseline results in the experiments section for a model that leverages intra-modal multi-scale information but fails to outperform AXM-Net. The proposed cross-modal cross-scale alignment is the key element of the AXM-Net, giving it an advantage over baseline and SOTA methods. The details of the layer-wise architecture of the feature learning backbone in the AXM-Net are provided in Table~\ref{architecture}.

\begin{figure*}[!htbp]
\centering
        \includegraphics[width=\linewidth,height=0.45 \textwidth]{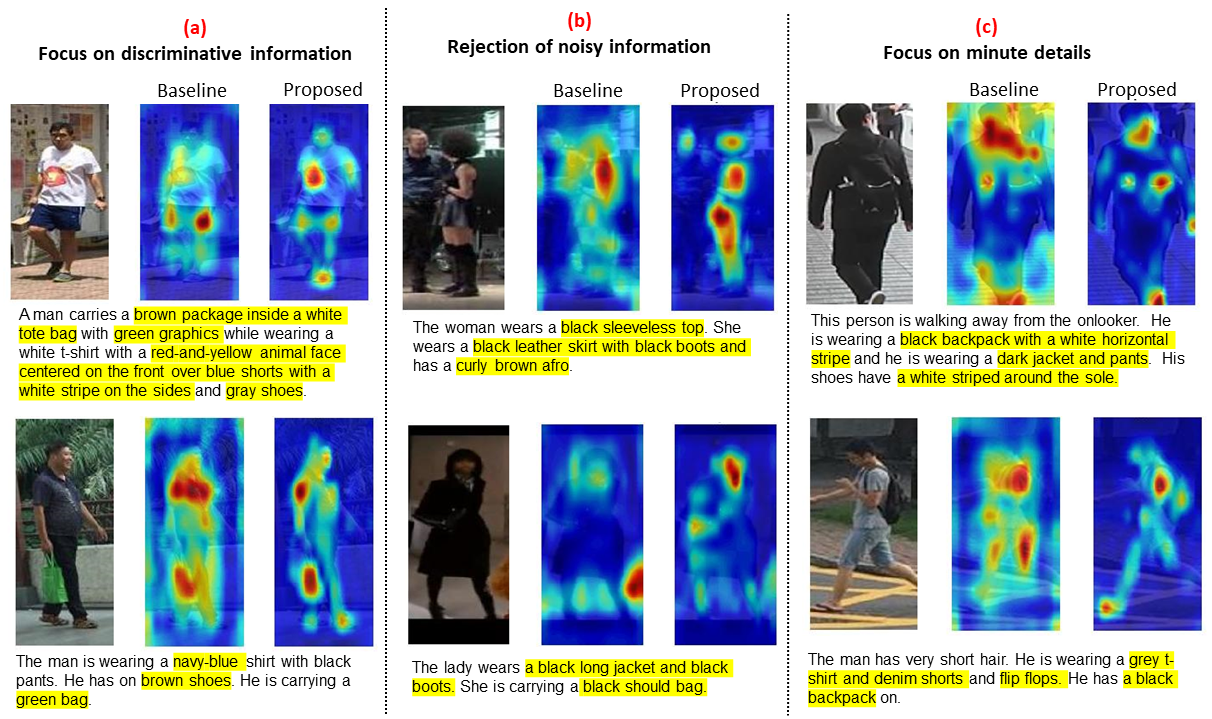}
        \caption{Visualisation of attention maps(warmer colours show higher attention). Baseline network and our proposed AXM-Net both leverage multi-scale features. High-lighted text phrases correspond to the semantic concepts of the person which are precisely attended by the proposed framework.}
    \label{attn_visual}
\end{figure*}

The implicit learning approach of AXM-Net utilises the textual input as a super-annotation signal and does not require auxiliary learning tasks such as segmentation, body landmark detection or word/phrase level matching, which is the case in the current SOTA methods~\cite{aggarwal2020text,wang2020vitaa, gumbel2020}. Since, the textual input does not have any background clutter, the effect of visual background is reduced. The alignment with textual feature during feature learning shown in Figure~\ref{attn_visual}.

\begingroup
\begin{table}[h!]
\centering
\resizebox{\linewidth}{!}{
\setlength{\tabcolsep}{1pt}
\begin{tabular}{c||c|c|c}
\hline
Stage & Module              & Vision Output & Textual Output \\ \hline \hline
input & - & 384 $\times$ 128, \ 3 & 1 $\times$ 56, \ 768 \\ \hline
conv1 & \begin{tabular}[c]{@{}c@{}}7 $\times$ 7 conv, stride=2, 3 $\times$  3 max pool \\ (1 $\times$ 3 conv, stride=1) on text\end{tabular} & 96 $\times$ 32, \ 64 & 1 $\times$ 56, \ 64  \\ \hline
Block1 & AXM block $\times$ 2 & 48 $\times$ 16, \ 256     &      1 $\times$ 28, \ 256\\ \hline
Block2 & AXM block $\times$ 2       & 24 $\times$ 8, \ 384 &   1 $\times$ 14, \ 384 \\ \hline
Block3 & AXM block $\times$ 2       & 24 $\times$ 8, \ 512 &   1 $\times$ 14, \ 512 \\ \hline
conv2 & 1 x 1 conv          & 24 $\times$ 8, \ 512 &   1 $\times$ 14, \ 512 \\ \hline
\end{tabular}}
\caption{Architecture of the unified feature learning network.}
\label{architecture}
\end{table}
\endgroup
\subsection{Visual Part-based Feature Coherence (VFC)}
The person Re-ID task depends on the discriminative local cues characterising each individual. The global visual branch focuses on the person as a whole while preserving shared semantics across modalities. However, it is important to pay attention to local cues and dependencies within each modality.

Recent method~\cite{wang2020img} used the sum of global channel-wise and spatial attention to attend intra-modal information. In channel attention, each channel is attended as a whole, while each pixel location is attended in spatial attention. In our case, we want to strengthen the relationship of semantic attributes to the associated body part as well as reinforce feature coherency across parts. To do this, we divide the image feature maps into $K$ horizontal strips $v_k$ where $k=1,2,\cdot\cdot, K$ . We apply max pool on each strip to get the most relevant features for a particular part. These features are then passed onto a MLP network to learn the relationship between features and their locality. The output ($p_k$) of the network attends each channel feature separately in each image part. Hence, it links the features to their location. For example, sneakers and laces are expected in foot region while a hat is expected on the head. This network is shared among all image strips to retain cross-part feature correspondence, like a long coat covers upper as well as lower body.
\begin{equation}
    {{p_k}} = \sigma(MLP(MaxPool({v_k})))
\end{equation}
\begin{equation}
    {{v_k}'} = p_{k} \odot v_{k}
\end{equation}
\begin{equation}
    {V_p} = conv_{1\times1}(concat({{v_k}}'))
\end{equation}
The attended image strips are concatenated back and fused into a single feature $V_{P}$ using linear layers for ID classification. The implicit feature alignment along with the VFC module offers a computational benefit by not requiring individual attribute/phrase level matching or multi-label learning.

\subsection{Self-Attention for Text (TSA)}
Similar to relationship among image parts, textual features are also required to model long-term dependencies among the phrases. To model these long-term dependencies we take inspiration from~\cite{wang2018non,cao2019gcnet} and employ a non-local self attention by directly computing interactions between the pairs of textual features. This step is important to complement our convolutional design for unified feature learning. We take feature maps from the last convolution block of the feature learning network as input to self-attention module. Intuitively, these features represent the important semantic attributes present in the person's description. Each spatial index represents a response from a local spatial region (phrases). The TSA module takes the feature from each spatial position, computes its affinity ($A$) with all other features(context) and attends input feature maps accordingly.

\subsection{Objective Function}
The loss function for training is the sum of cross-entropy (CE) loss of the three feature branches, triplet-loss~\cite{chechik2009large} and a simple affinity loss defined below. Specifically, the weights of the classifier layer are shared among all the branches to enforce intra-identity feature alignment. The CE loss is denoted as $\mathcal{L}_{ID_{joint}}$. The other losses are optimised in a pairwise manner for each visual and textural feature pair.
\setlength{\abovedisplayskip}{0pt} \setlength{\abovedisplayshortskip}{0pt}
\begin{multline}
    \mathcal{L}_{Total} = \lambda_{1} \mathcal{L}_{ID_{joint}} + \lambda_{2} [\mathcal{L}_{trip}(V_{g}, T) + \mathcal{L}_{trip}(V_{p}, T)] \\ + \lambda_{3} [\mathcal{L}_{aff}(V_{g}, T) + \mathcal{L}_{aff}(V_{p}, T)]
\end{multline}
where, $\lambda_i$, i=1,2,3 determines the proportion of each loss in the total.
Given tuples $(V_{a},T_{+}, T_{-})$, $(T_{a},V_{+}, V_{-}), \mathcal{L}_{trip}$ is a margin ($\alpha$) based ranking loss, defined over similarity($S$) of cross-modal positive and negative feature pairs as follows:
\setlength{\abovedisplayskip}{0pt} \setlength{\abovedisplayshortskip}{0pt}
\begin{multline}
    \mathcal{L}_{trip} = max[0,\alpha -(S(V_{a},T_{+})-(S(V_{a},T_{-})]\; + \\ 
    max[0,\alpha -(S(V_{+},T_{a})-(S(V_{-},T_{a})]
\end{multline}

\noindent\textbf{Cross-modal Affinity Loss:}
In order to enhance the retrieval performance, we propose to use a simple yet effective affinity loss. The image features and the corresponding textual features should be aligned and have high similarity, ideally +1 (in the case of cosine similarity) and non corresponding features should be uncorrelated and have low affinity. Given the representations for an image-text pair, affinity is measured in terms of the cosine similarity score between image feature $V$ and text feature $T$. The affinity loss is implemented as a binary cross entropy criterion, defined over $\{V_{i}, T_{j},y_{ij} \}$  where $y_{ij} = 1$ for matching pairs and $y_{ij} = 0$ for non-matching ones.
\setlength{\belowdisplayskip}{0pt} \setlength{\belowdisplayshortskip}{0pt}
\setlength{\abovedisplayskip}{0pt} \setlength{\abovedisplayshortskip}{0pt}
\begin{multline}
     \mathcal{L}_{aff} = -[\; y_{ij}\,.\,log(\sigma \ (S(V_{i},\ T_{j})))\,\\ + (1-y_{ij})\,.\,log(\sigma \ (1-S(V_{i},\ T_{j})))\; ]
\end{multline}
where $\sigma$ is sigmoid function applied to features similarity.

\begin{table*}[!h]
\centering
\resizebox{0.9\linewidth}{!}{
\begin{tabular}{l|c||cccc}
\hline
\multicolumn{1}{c|} {Model} & Feature Type  & Rank@1 & Rank@5 & Rank@10 & mAP \\ \hline \hline
GNA-RNN \cite{li2017person} & \multirow{7}{*}{global} &  19.05 & - & 53.64      & -    \\
IATV \cite{li2017identity}  &   &    25.94   & - & 60.48  &   -  \\
PWM \cite{chen2018Pwm} & &  27.14 & 49.45 & 61.02 & -\\
DPCE \cite{zheng2020dual}   &&  44.40 & 66.26 & 75.07& - \\
GLA \cite{chen2018improving} & &  43.58 & 66.93 & 76.26 & - \\
CMPC + CMPM \cite{zhang2018deep}  & &  49.37 &-& 79.27& - \\
\textbf{Baseline: Multi-scale features + joint ID} & &  52.78 & 72.33 & 80.29 & 49.04 \\
\hline
PMA \cite{jing2020pose} & global + keypoints &  53.81 & 73.54 & 81.23& - \\ \hline
ViTAA \cite{wang2020vitaa} & \multirow{2}{*}{global + attribute} &  55.97 &75.84 &83.52 & - \\
CMAAM \cite{aggarwal2020text}  &  & 56.68 & 77.18 & 84.86 & - \\
\hline
IMG-Net \cite{wang2020img} & \multirow{3}{*}{global + parts}  &  56.48 & 76.89 & 85.01 & - \\
HGAN \cite{gumbel2020} & & 59.00 & 79.49 & 86.68 & 37.80 \\
NAFS with RVN~\cite{gao2021contextual} & &  61.50 & 81.19 &  87.51 & - \\ \hline
\textbf{AXM-Net + joint ID} & \multirow{5}{*}{global+ part}&  62.08 & 79.84 & 86.27 & 57.09 \\
\textbf{AXM-Net + joint ID + affinity }& & 62.80  & 80.52 & 87.11& 57.72    \\
\textbf{AXM-Net + joint ID + triplet} && 63.72 & 80.84  & 87.16 & 58.46  \\
\textbf{AXM-Net + joint ID + triplet + affinity}  && \textbf{64.44} & 80.52 & 86.77 & 58.73 \\
\hline
\end{tabular}}
\caption{Comparison with SOTA models on the CUHK-PEDES dataset}
\label{SOTA}
\end{table*}

\section{Experiments and the Results}
\subsection{Implementation Details}
We follow a two stage training strategy to train the AXM-Net. In the first stage, we focus on learning the textual branch, the VFC branch and all fully connected layers from scratch, while keeping the vision backbone fixed to pretrained ImageNet weights. For the first stage the training follows the standard classification paradigm considering each person as an individual class and only using $\mathcal{L}_{ID_{joint}}$. We also apply label smoothing to our cross entropy loss. We use batch size 64, weight decay 5e-4 and initial learning rate 0.01 with stochastic gradient descent optimisation. Images are resized to 384 $\times$ 128. Each textual description is mapped to a 768 dimensional BERT embedding~\cite{devlin2018bert} and resized as 1 $\times$ 56 $\times$ 768 where 56 is the maximum sentence length. We kept the word embedding layer fixed during training.
We adopted random flipping, random erasing for images, and random circular shift of sentences as data augmentation. To achieve computational efficiency, we employ depth-wise separable convolutions at each layer. We used equal contribution($\lambda$) of each loss and margin ($\alpha$) equal to 0.5.
During inference, the vision and text features are extracted separately as the weights of the unified feature learning backbone are independent of data samples. We use the sum of both vision features ($V_{G} + V_{P}$) and text feature for evaluation. The performance is measured based on the cosine similarity between these features and reported in terms of Rank@1 and mean average precision ({m}AP).

\subsection{Datasets}
\noindent\textbf{CUHK-PEDES:}
The CUHK person description data \cite{li2017person} is the only large-scale benchmark available for cross-modal person search. It has 13003 person IDs with 40,206 images and 80,440 descriptions. There are 11003,1000,1000 pre-defined IDs for training, validation and test sets. The training and test set include 34054/68126 and 3074/6156 images/descriptions respectively. 
\noindent\textbf{CrossRe-ID Dataset:}
For cross-modal Re-ID, we evaluate the models on the protocol introduced by \cite{farooq2020IJCB} on the test split of CUHK-PEDES data. The gallery and query splits have been carefully separated across viewpoints. The descriptions are also varying across viewpoints. The dataset includes 824 unique IDs. There are 1511/3022 and 1096/2200 images/descriptions in gallery and query sets respectively. 
\noindent\textbf{CUHK-SYSU:}
We evaluate our model on the test protocol provided by \cite{farooq2020convolutional}. 
There are 5532 IDs for training and 2099 IDs for testing. The corresponding descriptions have been extracted from CUHK-PEDES data. The final gallery and query splits contain 5070/10140 and 3271/6550 images/descriptions respectively.

\subsection{Comparison with State-of-the-Art Methods}
\subsubsection{Results on Person Search}
We summarise the performance of the proposed AXM-Net with the SOTA methods on the CUHK-PEDES data in Table ~\ref{SOTA}. The methods have been grouped according to the type of representations used for learning. We implement the baseline network with multi-scale features for both modalities and a joint classifier layer. The baseline method performs best among all global level techniques which signifies the benefit of having multi-scale features but still it falls behind the other complex methods~\cite{wang2020vitaa,aggarwal2020text,wang2020img,gumbel2020} due to lack of cross-modal alignment.
It is worth noting that our method is simple but powerful and enhances the semantic alignment between modalities without any explicit supervision from segmented body parts~\cite{wang2020vitaa}, attributes~\cite{aggarwal2020text} and phrases/words~\cite{gumbel2020}. The proposed AXM-Net with simple $\mathcal{L}_{ID_{joint}}$ loss outperforms current SOTA by a large margin, achieving over 62\% Rank@1 performance. By using affinity and triplet loss together, we set the new SOTA Rank@1 of 64.44\%.

\begingroup
\begin{table*}[]
\resizebox{\linewidth}{!}{
\setlength{\tabcolsep}{2pt}
\begin{tabular}{l|cc|cc|cc||cc|cc|cc}
\hline
\multicolumn{1}{c|}{\multirow{3}{*}{Model}} & \multicolumn{6}{c||}{\textbf{CrossRe-ID}}                  & \multicolumn{6}{c}{\textbf{CUHK-SYSU}}  \\ \cline{2-13} 
\multicolumn{1}{c|}{} &
  \multicolumn{2}{c|}{V $\xrightarrow{}$ V} &
  \multicolumn{2}{c|}{T $\xrightarrow{}$ V} &
  \multicolumn{2}{c||}{VT $\xrightarrow{}$ V} &
  \multicolumn{2}{c|}{V $\xrightarrow{}$ V} &
  \multicolumn{2}{c|}{T $\xrightarrow{}$ V} &
  \multicolumn{2}{c}{VT $\xrightarrow{}$ V} \\ \cline{2-13} 
\multicolumn{1}{c|}{}                       & Rank@1 & mAP   & Rank@1 & mAP   & Rank@1 & mAP   & Rank@1 & mAP   & Rank@1 & mAP   & Rank@1 & mAP   \\ \hline \hline
JT + CCA~\cite{farooq2020IJCB} & 86.77  & 88.90 & 33.61  & 39.40 & 88.59  & 87.95 &  74.13 & 77.16 & 11.37 & 15.78 & 77.68 & 75.8\\
AXM-Net + joint ID + affinity               & 95.14  & 96.04 & 44.66  & 50.49 & 95.26  & 95.22 & 86.00 & 87.75 & 19.93 & 24.82 & 88.72 & 87.02\\
AXM-Net + joint ID + triplet & 95.02  & 96.00 & 47.33  & 52.58 & 95.75  & 95.41 & 86.24 & 88.02 & 20.93 & 26.04 & 87.86 & 86.40 \\
AXM-Net + joint ID + affinity + triplet     & 94.29  & 98.9  & 46.48  & 52.21 & 94.05  & 93.93 & 85.86 & 87.70  & 21.44 & 26.77  & 88.62 & 86.73 \\ \hline
\end{tabular}}
\caption{Performance comparison on cross-modal Re-ID. Query $\xrightarrow{}$ Gallery}
\label{crossModal}
\end{table*}
\endgroup

\begin{table}[]
\center 
\resizebox{\linewidth}{!}{
\begin{tabular}{c|ccc||c}
\hline
\textbf{Model} & \textbf{AXM-Block} & \textbf{VFC} & \textbf{TSA} & \textbf{Rank@1} \\ \hline
Baseline & - & - & -   & 52.78  \\
Model: 1 & \checkmark& - &  -  & 55.90 \\
Model: 2 & - & \checkmark&  -   & 56.99 \\
Model: 3 & - & - &  \checkmark  & 53.25 \\ \hline
Model: 4 & \checkmark& - & \checkmark  & 56.27 \\
Model: 5 & \checkmark& \checkmark&  -   & 58.59 \\ \hline
Unified Visual   & \checkmark& \checkmark&  \checkmark & 54.53   \\
Single Stage & \checkmark& \checkmark&  \checkmark & 55.05 \\
AXM-Net & \checkmark& \checkmark&  \checkmark & 59.44\\
\hline
\end{tabular}}
\caption{Ablation study on the AXM-Net on CUHK-PEDES test set with Word2vec text embedding and $224 \times 224$ image size}
\label{ablation}
\end{table}

\subsubsection{Results on Cross-modal Re-ID}
We follow the evaluation protocol of ~\cite{farooq2020IJCB} for cross-modal re-identification. For fair-comparison, we used image size 224$\times$224 and word2vec~\cite{mikolov2013distributed} embedding. In Tables \ref{crossModal}, V $\xrightarrow{}$ V indicates image based search, T $\xrightarrow{}$ V indicates description to image search and VT $\xrightarrow{}$ V indicates using both modalities for query and vision as gallery. We report the detailed results including Rank@5 and Rank@10 for all the datasets in the supplementary materials.
For CrossRe-ID and CUHK-SYSU, we compare the results with the recently reported work ~\cite{farooq2020IJCB}. It is mentioned as JT+CCA in Tables \ref{crossModal}. For both datasets, the proposed AXM-Net outperformed the previous method by a significant margin in all retrieval scenarios. Note that, now the  T $\xrightarrow{}$ V indicates a description of a person from a different pose, and the gallery images have different poses. We can witness the potential of semantic alignment across-modalities in this challenging case. The improvement in Rank@1 performance shows the capability of the AXM-Net in matching viewpoint-invariant semantic details.

\subsection{Component Analysis}
We perform an ablation study for the proposed framework by adding the components step-by-step. The study is performed on the CUHK-PEDES test set with joint cross-entropy loss $\mathcal{L}_{ID_{joint}}$. All hyper-parameters are kept the same for training in all settings. Table ~\ref{ablation} presents the corresponding results. The \textbf{baseline} model has the same architecture as the global visual and textual branch, including multi-scale features for both modalities but without alignment. We list our observations as follows:

\begin{itemize}
    \item \textbf{Effect of Individual Components.} \textbf{Models:\ 1,\ 2,\ 3} correspond to the contributions of individual components. We note that each component has boosted the retrieval capability compared to the baseline. Each component is essential to the design of cross-modal Re-ID. The semantic alignment induced by the unified feature learning enhances the useful information across modalities. The VFC branch attends the locally informative cues, while the TSA keeps track of textual dependencies. \textbf{Models:\ 4,\ 5} also emphasise the complementary effect of various components together.
    \item \textbf{The Effect of the Unified Visual Feature Branch.} 
    We experiment with the unified visual branch by removing the global branch and keeping only the VFC based part branch. It is indicated as \textbf{Unified Visual} in Table \ref{ablation}. We observe a performance drop of 4.58\%, as compared to the proposed design. It is important to extract the global feature and perform a local enhancement separately to retain the cross-modal alignment learnt by the backbone network.
    \item \textbf{The effect of Single Stage versus Two Stage  Learning.}
    As mentioned earlier, we used a two stage training for network learning. We also test a single stage policy in which we tune all the parameters together with the same initialisation setting. \textbf{The single stage} model clearly shows the difference between the two policies. Hence, it is recommended to learn weakly aligned textual features beforehand to support best convergence. 
\begin{table}[]
\center 
\resizebox{\linewidth}{!}{
\begin{tabular}{c|c||c|c}
\hline
MLP Weights & Rank@1 & Pooling Type     & Rank@1 \\ \hline
Separate          & 57.62  & Average (GAP)         & 57.36  \\
Shared            & 57.93  & Max (GMP)             & 58.82  \\ \hline\hline

Feature Drop      & Rank@1 & No. of FC Layers & Rank@1 \\ \hline
Random            & 57.62  & 1                & 58.45  \\
Part              & 58.33  & 2                & 58.06  \\
No drop           & 59.44  & Proposed         & 59.44  \\ \hline
\end{tabular}}
\caption{Design parameters for the VFC branch}
\label{part-design}
\end{table}

    \item \textbf{Design Parameters for the VFC Branch.}
    We consider several parameters for designing the part-based visual feature branch as presented in Table \ref{part-design}. First of all, we test with separate MLP networks for each strip of image features. We find that a shared network helps feature coherency as well as reducing the number of parameters. It supports our idea of context sharing between image parts and signifies the connectedness of various semantic concepts across the strips. Next, we note that the global max pool helps in capturing local cues by focusing on the highest responses in each region. For each branch of AXM-Net, we also optimise the number of FC layers as it is critical to avoid any performance degradation and network overfitting. We observe from the table that having an identical linear layer structure not always implies the best performing solution. Being a richer modality and focusing on information from the whole image, two FC layers help in the global branch to learn better representations. We also consider the feature dropping technique~\cite{dai2019batch} to obtain robust features. However, it did not help either by random batch drop block or horizontal strip (part) drop. 
    \item \textbf{Position of Alignment} In Table~\ref{axm-position}, we investigate the effect of cross-modal alignment in different stages. The best performance is achieved by aligning at all levels. The semantic concepts at lower layers are not well defined but still adds a little benefit in our case. It is due to the fact that the first strided (7$\times$7) convolution on image captures rich set of primary features, also practised in modern vision Transformers. Hence, the lower blocks also get significant contextual information of the image. 
\begin{table}[!htbp]
\center 
\resizebox{\linewidth}{!}{
\begin{tabular}{c|ccc|c}
\hline
 & Block 1 & Block 2 & Block 3 & Rank@1 \\ \hline 
\multirow{5}{0.25\linewidth}{\centering  Position of AXM-Block}        &      -   &  -       &  -& 58.41      \\
  & \checkmark    &  \checkmark       & -    & 59.16       \\
  &  \checkmark    &     -    & \checkmark        & 59.24        \\
   & -   &     \checkmark      &   \checkmark      & 59.41     \\
     &\checkmark   &   \checkmark      &    \checkmark     &   59.44     \\ \hline
\end{tabular}}
\caption{Effect of implicit alignment in different blocks.}
\label{axm-position}
\end{table}

    \item \textbf{The Effect of Language Embedding.} We notice in Table~\ref{tab:wordEmbed} that the performance of the system can be improved by using complex and richer language embeddings. We hope to make further gains by combining language modelling with cross-modal feature learning.
    \end{itemize}

\begin{table}[!htbp]
\centering
\resizebox{\linewidth}{!}{
\begin{tabular}{l|cc}
\hline
\multicolumn{1}{c|}{Rank@1}             & Word2Vec & BERT  \\ \hline
AXM-Net + joint ID                      & 59.44    & 61.27 \\ \hline
AXM-Net + joint ID + triplet + affinity & 61.9     & 63.0  \\ \hline
\end{tabular}}
\caption{Effect of choice of language embedding.}
\label{tab:wordEmbed}
\end{table}

\section{Conclusion}

We presented a novel AXM-Net model to address the challenges of cross-modal person re-identification. Our innovation involves: 1) a unified feature learning block, called AXM-Block, which implicitly aligns the semantic features across the visual and textual modalities, and 2) an effective design for reinforcing feature coherence among image parts. In contrast to existing methods, the proposed AXM-Net is the first framework that is based on an integrated cross-modal feature alignment and learning stage in Re-ID context. An important advantage of the proposed method is that it does not rely on external cues for explicit alignment of feature embeddings. The experimental results show that our network defines new SOTA performance on the CUHK-PEDES benchmark and also demonstrates the potential of the proposed network for more challenging cross-modal person Re-ID applications.

\section{Acknowledgements}
This work has been supported in part by the EPSRC Grants; FACER2VM (EP/N007743/1), MVSE (EP/V002856/1), JADE2 (EP/T022205/1), and the EPSRC/dstl/MURI project EP/R018456/1.

\bibliography{egbib.bib}

\end{document}